# A Lemma Based Evaluator for Semitic Language Text Summarization Systems

Tarek El-Shishtawy[1] and Fatma El-Ghannam[2]

[1] Benha University
Cairo, Faculty of Engineering (Shoubra) , Egypt

[2] Electronics Research Institute
Cairo, Egypt

**Abstract**

Matching texts in highly inflected languages such as Arabic by simple stemming strategy is unlikely to perform well. In this paper, we present a strategy for automatic text matching technique for for inflectional languages, using Arabic as the test case. The system is an extension of ROUGE test in which texts are matched on token's lemma level. The experimental results show an enhancement of detecting similarities between different sentences having same semantics but written in different lexical forms..

## 1. Introduction

Evaluation of automatic text summarization systems by human evaluators requires massive efforts. This hard expensive effort has held up researchers looking for methods to evaluate summaries automatically. Current automated methods compare fragments of the summary to be assessed against one or more reference summaries (typically produced by humans), measuring how much fragments in the reference summary is present in the generated summary.

In Semitic languages such as Arabic, many sentence structures can be written to express the same semantic. This adds more complexity to assessed tools that measures the similarity of human and machine summaries. Therefore, in this paper, we propose a lemma based system for evaluating similarity of given texts..

The presented work is an extension of ROUGE test, which is widely used for evaluating the performance of summarization techniques. The presented work is based on representing words in their lemma forms, and modifying text similarity to be evaluated on 'token' instead of 'character' level. The results show that this technique is more suitable for measuring similarity of Arabic texts.

The remainder of this paper is organized as follows: section 2 describes ROUGE evaluation measure and other previous works. In section 3, we discuss the lemma based form of Arabic words. Section 4 describes the proposed lemma based Rouge evaluation technique. Section 5 describes the datasets and the experiment used to evaluate the proposed system.

## 2. Previous Works

Evaluation of automatic text summarization systems can be extrinsic or intrinsic evaluation methods. In extrinsic evaluation, the summary quality is judged on the basis of how helpful summaries are for a given task, such as how well they can answer certain questions relative to the full source text. Intrinsic evaluation has mainly focused on the informativeness of summaries. This is often done by comparing the system summary to reference/human summary. ROUGE [8] stands for Recall-Oriented Understudy for Gisting Evaluation is a recall measure that counts the number of overlapping n-gram[1] units between the system summary generated by computer and several reference summaries produced by human. Rouge has proved to be a successful algorithm [9]. Several variants of the measure were introduced, such as Rouge -N, and Rouge -S. Rouge-N is an n-gram recall between a candidate summary and a set of reference summaries. It is defined by the following formula:

$$ROUGE-N = \frac{\sum_{s \in (reference summaries)} \sum_{gram_n \in s} count_{match}(gram_n)}{\sum_{s \in (reference summaries)} \sum_{gram_n \in s} count(gram_n)}$$

(1)

where n is the length of the n-gram; *$gram_n$*, *$count(gram_n)$* is the number of n-grams in the reference summaries, and *$count_{matches}(gram_n)$* is the maximum number of n-grams co-occurring in a system summary and in reference summaries.

Rouge-S utilizes Skip-Bigram Co-Occurrence Statistics. Skip-bigram any pair of words in their sentence order, allowing for arbitrary gaps. Rouge-S, measures the overlap ratio of skip-bigrams between a candidate summary and a set of reference summaries. In our work, we adopted a lemma based Rouge measure by applying an Arabic lemmatizer [5] for stemming in the Rouge evaluation process.

---

[1] An n-gram is a subsequence of n words from a given text.

## 2. Lemmatization of Arabic Words

There is no general agreement of the representation level of Arabic words in IR systems. Two levels have been debated; root level and stem level. Motivated by the power of Arabic roots, the first approach represents words in their root forms. Historically, a root was the entry to traditional Arabic lexicons. Almost, every word in Arabic is originated from a root. Many researchers [10, 11], have emphasized that the use of Arabic roots as indexing terms substantially improves the information retrieval effectiveness. Several researchers [1,3] criticized this approach, and based their representation on stems. A stem is the least marked form of a word that is the uninflected word without suffixes, prefixes.

The main problem in selecting a root as a standard representation level in information retrieval systems is the over-semantic classification. Many words that do not have similar semantic interpretations are grouped into same root. On the other hand, stem level suffers from under-semantic classification. Stem pattern may exclude many similar words sharing the same semantic properties.

In this work, we use the lemma form of Arabic words. In the field of NLP, lemmatization refers to the process of relating a given textual item to the actual lexical or grammatical morpheme [5]. Lemma refers to the set of all word forms that have the same meaning, and hence capture semantic similarities between words. El-Shishtawy and El-Ghannam [7] proposed the first non-statistical accurate Arabic lemmatizer algorithm that is suitable for information retrieval (IR) systems. The proposed lemmatizer makes use of different Arabic language knowledge resources to generate accurate lemma form and its relevant features that support IR purposes.

## 4. Lemma Based Evaluation

We have modified the existing Rouge evaluation methods implemented in the Dragon Toolkit Java based package (http://dragon.ischool.drexel.edu/ ). Rouge is based on counting the number of n-grams co-occurring in a candidate summary and a set of reference summaries. Human judge can match different word forms presented by the machine and a reference summaries such as: *organize, organizes, and organizing*. Stemming has a large effect on Arabic information retrieval, due to the highly inflected nature of the language. Rouge test can accept different word granularity level of the n-grams terms (original word or stem form). The Dragon toolkit supports the extraction of word stem by the Porter stemmer, but does not support the Arabic language. As illustrated in section 2, the lemma form can relate different word forms that have the same meaning. The main advantage of lemma form is that it can conflate different word forms that stem can not do. Therefore, The first modification is to apply the Arabic lemmatizer for extracting the word canonical forms for both documents to be compared, prior to the Rouge evaluation process.

The original ROUGE is an n-grams system where grams are set in terms of number of characters. This may be useful when comparing non lemmatized words hoping to overcome morphological variations. The used lemma forms eliminate the need for character level comparisons. It was shown in other works that text similarity measurement can be greatly enhanced when measured at token's level [8, 9]. . Therefore, the second modification is to use 'tokens' instead of 'characters' as a basis of comparison. The modifications are given by the following formula:

$$LmmaROUGE-N = \frac{\sum_{s \in (referencesummaries)} \sum_{wordlemma_n \in s} count_{match}(WordLemma_n)}{\sum_{s \in (referencesummaries)} \sum_{Word_n \in s} count(Word_n)} \quad (2)$$

To perform the test, first the machine and reference summaries are applied to the Arabic lemmatizer to extract their lemma forms. Second, the output from the lemmatizer is passed to the Dragon toolkit for the evaluation process with grams modified to be words. We apply an experiment to compare the Rouge results with and without using the lemmatizer.

**5. Datasets and Results**

To test the proposed lemma based evaluator, two datasets were adopted. The first (DataSet1) was TAC 2011 Dataset[2]. It was selected to evaluate and compare the summaries of the new techniques against the published summaries of systems presented in TAC 2011, using the same dataset for the same task of producing 240-250 words summary. DataaSet1 included the source texts, system summaries, and human summaries. The data set is available in 7 languages including Arabic. It was derived from publicly available WikiNews English texts. Texts in other languages have been translated by native speakers of each language. The source texts contain ten collections of related newswire and newspaper articles. Each collection contains a cluster of ten related articles. The average number of words per article is 235 words. Each cluster of related articles deals with news about a single event. Each article in a certain cluster includes a limited number of topics (mostly one or two topics) about the event. Some topics are dealt with by many documents, while others are addressed in a single document.

---

[2] (http://www.nist.gov/tac/2011/Summarization/).

## 6 Experimental Results

Arabic summaries of the dataset described in the previous section were used in this experiment. The dataset already includes the summarization results generated by systems under test as well as the reference human summaries. The proposed lemma based evaluator was applied to all summaries, and the experimental results were reported in terms of Rouge-2 and Rouge-S. Precision P, recall R, and F-measure were calculated for each measure.

Tables (1,2) show a comparison between the evaluation techniques when applied to different existing summarization systems. Evaluation techniques are native Rouge-2 and native Rouge-S measures, and proposed lemma Rouge measures. The results show that the two proposed techniques perform well when compared to native ROUGE/. In all evaluations, Sen-Rich technique outperformed all systems by a significant margin in Rouge-S measure; it gains 0.2014, compared to the next ranking 0.1585 for Classy and 0.1554 for Doc-Rich.

Another experiment was also applied to compare the effect of adding the lemmatizer in the Rouge test. We applied the Rouge test without adding the lemmatizer using the Dragon toolkit, and comparing the results with that previously reported in the case of adding the lemmatizer. As expected, the results show better results for all systems when using the lemmatizer. This improvement is due to increasing the number of matching tokens between the machine and reference summaries when using the lemmatizer. The proposed system was used also in evaluating multiple-documents summarization [4, 6].

*Table 3  A comparison between the lemma-based technique for evaluating different systems using Rouge-2*

| System | Lemma Based Rouge2 F-Measure | Native Rouge2 F-Measure |
|---|---|---|
| CLASSY1 | 0.1529 | 0.1442 |
| TALN_UPF1 | 0.1326 | 0.1099 |
| CIST1 | 0.1279 | 0.0871 |
| UoEssex1 | 0.1165 | 0.1129 |
| UBSummarizer1 | 0.0915 | 0.0711 |

*Table 4  A comparison between the lemma-based technique for evaluating different systems using Rouge-S*

| System | Lemma Based RougeS F-Measure | Native RougeS F-Measure |
|---|---|---|
| CLASSY | 0.1585 | 0.1302 |
| TALN_UPF1 | 0.1472 | 0.1213 |
| UoEssex1 | 0.1331 | 0.1123 |
| CIST1 | 0.0973 | 0.0786 |
| UBSummarizer1 | 0.0838 | 0.0640 |

**Conclusion**

We have presented a technique for measuring similarity between texts suitable for Semitic languages. The highly inflections inherent to such languages -such as Arabic- makes the current evaluation techniques not suitable for detecting semantically equivalent sentences written in different syntactic forms.  We extended ROUGE system so that texts are matched on token's lemma level. The published results of TAC2011 systems were used to evaluate different systems using the new technique. The experimental results indicated that adding the lemmatizer improves the Rouge results of all systems.